\pdfoutput=1

\documentclass[11pt]{article}

\usepackage{ACL2023}

\usepackage{times}
\usepackage{latexsym}

\usepackage[T1]{fontenc}

\usepackage[utf8]{inputenc}

\usepackage{microtype}

\usepackage{inconsolata}

\usepackage{graphicx}
\usepackage{subcaption}
\usepackage{amsmath}
\usepackage{stackengine}
\usepackage{tabularx}
\usepackage{multirow}
\usepackage{booktabs}

\title{FOAL: Fine-grained Contrastive Learning for Cross-domain Aspect Sentiment Triplet Extraction}

\author{Ting Xu\textsuperscript{\rm $\spadesuit$}\;,
 Zhen Wu\textsuperscript{\rm $\spadesuit$}\;, 
 Huiyun Yang\textsuperscript{\rm $\clubsuit$}, 
 Xinyu Dai\textsuperscript{\rm $\spadesuit$}\\
\textsuperscript{\rm $\spadesuit$}National Key Laboratory for Novel Software Technology, Nanjing University\\
\textsuperscript{\rm $\clubsuit$}ByteDance\\
\texttt{xut@smail.nju.edu.cn}, \texttt{\{wuz, daixinyu\}@nju.edu.cn}\\
\texttt{yanghuiyun.11@bytedance.com}
}
\begin{document}
\maketitle
\newcommand{\method}{\textsc{FOAL }}
\newcommand{\methodwb}{\textsc{FOAL}}
\begin{abstract}
Aspect Sentiment Triplet Extraction (ASTE) has achieved promising results while relying on sufficient annotation data in a specific domain. However, it is infeasible to annotate data for each individual domain. We propose to explore ASTE in the cross-domain setting, which transfers knowledge from a resource-rich source domain to a resource-poor target domain, thereby alleviating the reliance on labeled data in the target domain. To effectively transfer the knowledge across domains and extract the sentiment triplets accurately, we propose a method named Fine-grained cOntrAstive Learning (\methodwb) to reduce the domain discrepancy and preserve the discriminability of each category.
Experiments on six transfer pairs show that \method achieves 6\% performance gains and reduces the domain discrepancy significantly compared with strong baselines. Our code will be publicly available once accepted.
\end{abstract}
\begin{figure*}
    \centering
    \includegraphics[width=\textwidth]{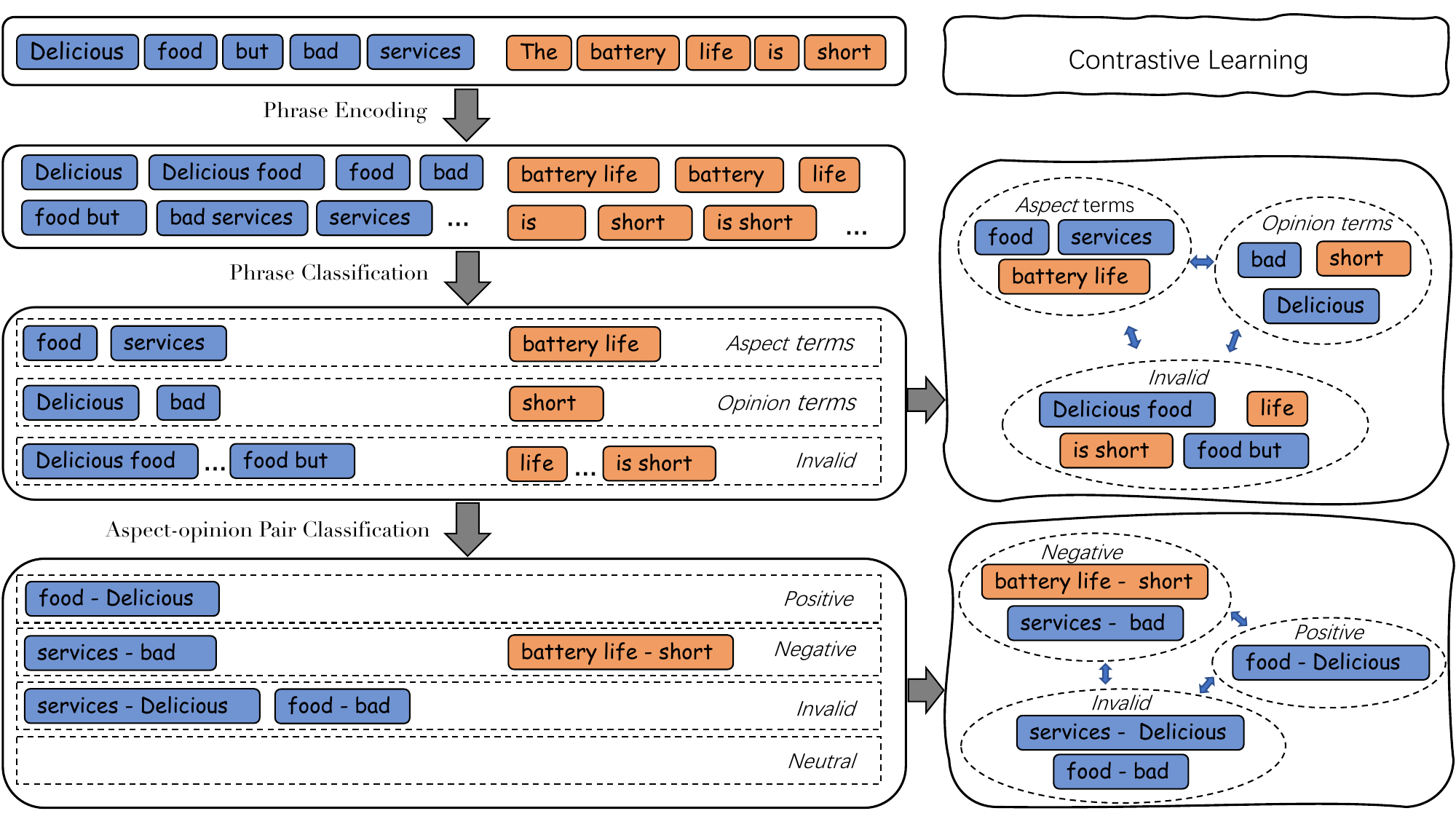}
    \caption{An overview of \method  architecture. The blocks represented in blue and orange correspond to features from the source and target domains, respectively. We utilize golden labels from the source domain and pseudo-labels from the target domain. In contrastive learning, positive pairs are constructed using phrase/pair features from the same category and negative pairs are constructed from different categories. 
    Both the positive and negative pairs are across different domains. Fine-grained contrastive learning can reduce the domain discrepancy while preserving the discriminability of each category.
    }
    \label{fig:model}
\end{figure*}
\section{Introduction}
Aspect Sentiment Triplet Extraction \citep[ASTE;][]{peng2020knowing}, a fine-grained sentiment analysis \citep{medhat2014sentiment} task, has attracted considerable interest recently. It focuses on extracting the sentiment triplets from a given review, ie., aspect terms, opinion terms, and sentiment polarity. Prior works \citep{xu-etal-2021-learning, yan-etal-2021-unified} on ASTE have achieved promising results while heavily relying on massive annotations in a specific domain.

However, in practice, customer reviews can originate from a wide range of domains (e.g. Amazon Review \citep{ni-etal-2019-justifying} covers 29 domains). Consequently, it may be infeasible to annotate a sufficient amount of data for each individual domain. To address this issue, we propose to explore ASTE in the cross-domain setting, which transfers knowledge from a resource-rich source domain to a resource-poor target domain, thereby reducing the reliance on labeled data in the target domain. 

The task of cross-domain ASTE presents unique challenges in terms of transferability and discriminability. From a \textbf{transferability} perspective, the model trained from a source domain is hindered by the variability of language used across domains. The terminology and phraseology used in different domains differ a lot, making it difficult for models to understand the meaning in a new target domain. \textbf{Discriminability}, in the context of ASTE, refers to the ability of the model to accurately identify aspect terms, opinion terms, and sentiments within the target domain. In other words, it is the capability of a model to discern between aspect terms and non-aspect terms, as well as between opinion terms and non-opinion terms, and to classify sentiments accordingly. For example, given the text "The battery life is short", a model with high discriminability should be able to correctly identify "battery life" as the aspect term, "short" as the opinion term, and negative as the sentiment. To summarize, cross-domain ASTE requires the transfer of knowledge and accurate extraction of sentiment triplets across different domains.

To address the above challenges, we propose a new domain adaptation strategy, named Fine-grained cOntrAstive Learning (\methodwb), which utilizes contrastive learning \citep{jaiswal2020survey} to learn domain-invariant representations while preserving the discriminability of the fine-grained factors in ASTE, i.e., aspect terms, opinion terms, and sentiments. Specifically, we select two features, one from the source domain and the other from the target domain, to construct positive or negative pairs. \textit{The positive pair has the same category, while the negative pair has a different category}. By pulling the positive pairs together, we can  reduce the discrepancy between domains in the corresponding categories, thereby improving the transferability of the model. By pushing the negative pairs apart, we can better discern between aspect terms and non-aspect terms, as well as between opinion terms and non-opinion terms, and to classify sentiments accordingly, thereby improving discriminability.

We evaluate \method on six transfer pairs of \citet{xu-etal-2020-position}. Results show that \method outperforms the baseline model by 6\% in F1 scores. And quantitative analysis demonstrates that \method can reduce the domain discrepancy while preserving the category discriminability.
\section{Methodology}
\subsection{Backbone Model}
Our backbone model is Span-ASTE \citep{xu-etal-2021-learning}, a representative model in ASTE \footnote{We introduce this method briefly, refer to \citet{xu-etal-2021-learning} for more details.}. 
Given a text with $n$ tokens $w_1, w_2,...,w_n$, it first obtains the token, phrase, and pair representations corresponding to the word, phrase, and aspect-opinion pair, respectively:
\begin{equation}
    \begin{split}
        [\mathbf{h}_1, \mathbf{h}_2...,\mathbf{h}_n] &= \mbox{BERT}(w_1, w_2,..., w_n),\\
        \mathbf{s}_{i,j} &= [\mathbf{h}_i; \mathbf{h}_j; f_{width}(i, j)],\\
        \mathbf{g}_{\mathbf{s}_{b,c}^a, \mathbf{s}_{d,e}^o} &= [\mathbf{s}_{b, c}^a; \mathbf{s}_{d,e}^o; f_{distance}(b, c, d, e)],\\
    \end{split}
    \label{eq:aste-rep}
\end{equation}
$\mathbf{s}_{i,j} \in \text{SP}$, where $i$ and $j$ are the start and end indices for the phrase $\mathbf{s}_{i,j}$,
and $\text{SP}$ is the set of all candidate phrases. Each phrase can be an aspect, opinion, or an invalid phrase. We conduct Cartesian Product \citep{agesen1995cartesian} on the candidate aspects and opinions to obtain pair representation $\mathbf{g}_{\mathbf{s}_{b,c}^a, \mathbf{s}_{d,e}^o}\in \text{PAIR}$, where $\text{PAIR}$ is the set of all candidate aspect-opinion pairs.
$f_{width}(i, j)$ and $f_{distance}(b, c, d, e)$ are two embedding layers for the phrases and pairs.
Then we employ classifiers to get scores for phrase type $m\in \{Aspect, Opinion, Invalid\}$ and pair type $r\in \{Positive, Negative, Neutral, Invalid\}$:
\begin{equation}
    \begin{split}
    &P(m|\mathbf{s}_{i, j})=\mbox{softmax}(\mbox{SPAN\_FFN}(\mathbf{s}_{i, j})),\\
    &P(r|\mathbf{s}_{b, c}^a, \mathbf{s}_{d, e}^o) = \text{softmax}(\text{PAIR\_FFN}(\mathbf{g}_{\mathbf{s}_{b,c}^a, \mathbf{s}_{d,e}^o})).
    \end{split}
    \label{eq:aste-prob}
\end{equation}
The training loss is defined as the sum of the negative log-likelihood for the phrase and pair scores:
\begin{equation}
    \begin{split}
        \mathcal{L}_{aste} = &- \sum_{\mathbf{s}_{i,j}\in \text{SP}} \log P(m=m_{i,j}^*|\mathbf{s}_{i,j}) \\
        &- \sum_{(\mathbf{s}_{b, c}^a, \mathbf{s}_{d, e}^o) \in \text{PAIR}}\log P(r=r^*|\mathbf{s}_{b, c}^a, \mathbf{s}_{d, e}^o),
    \end{split}
    \label{eq:aste-loss}
\end{equation}
where $m^*_{i,j}$ and $r^*$ are the golden labels for phrase $\mathbf{s}_{i,j}$ and the aspect-opinion pair $(\mathbf{s}_{b, c}^a, \mathbf{s}_{d, e}^o)$.

\subsection{Fine-grained Contrastive Learning}
\paragraph{Basic Intuition.} Our ultimate goal is to transfer knowledge across domains and accurately extract the sentiment triplets in the target domain. This requires us to remain the discriminability of different categories while reducing the domain discrepancy across domains. To achieve this goal, we select two features, one from the source domain and the other from the target domain, to construct positive or negative pairs. The positive pair has the same category, while the negative pair has a different category. 
By pulling the positive pairs together, we can reduce the discrepancy across domains in the corresponding category. By pushing the negative pairs apart, we can better distinguish aspect terms from non-aspect terms, opinion terms from non-opinion terms, and different sentiments.

\paragraph{Positive and Negative Pairs.}  We draw the positive and negative pairs from both the source and target domains. Since there is no annotation for the target domain, we use the pseudo label predicted by the backbone model, then pull features of the same type together and push features of different types away. Specifically, we employ an indicator function $c(\mathbf{x}_i, \mathbf{x}_j, t)$ for a feature $\mathbf{x}_i$ from the source domain and a feature $\mathbf{x}_j$ from the target domain. If $\mathbf{x}_j$ has the same prediction as $\mathbf{x}_i$ with the prediction score higher than $t$, they are positive pairs ($c(\mathbf{x}_i, \mathbf{x}_j, t)$=1). Otherwise, they are negative pairs ($c(\mathbf{x}_i, \mathbf{x}_j, t)$=0).  For example in Figure \ref{fig:model}, for "food" in the source domain, "battery life" in the target domain is the positive example, and all other non-aspect terms in the target domain are the negative examples.

\paragraph{Contrastive Learning.} Given the positive and negative pairs, we can obtain contrastive loss from the source-target and target-source directions:
\begin{equation}
\small
\begin{split}
    \mathcal{L}_{contra}(S, T, t) &= -\sum_{\mathbf{x}_i \in S} \sum_{\mathbf{x}_j\in T}\log\frac{d(\mathbf{x}_i, \mathbf{x}_j)c(\mathbf{x}_i, \mathbf{x}_j, t)}{\sum_{\mathbf{x}_k \in T}d(\mathbf{x}_i, \mathbf{x}_k)}\\
    &-\sum_{\mathbf{x}_i \in T} \sum_{\mathbf{x}_j\in S}\log\frac{d(\mathbf{x}_i, \mathbf{x}_j)c(\mathbf{x}_j, \mathbf{x}_i, t)}{\sum_{\mathbf{x}_k \in S}d(\mathbf{x}_i, \mathbf{x}_k)},\\\\
    d(\mathbf{x}_i, \mathbf{x}_j) &= \exp(\cos(\mathbf{x}_i, \mathbf{x}_j)/\tau),\\
\end{split}
\label{eq:contra-abs}
\end{equation}
where $S$ and $T$ are two feature sets from the source and target domains, $\tau$ is the temperature hyper-parameter, $\exp$ denotes the exponential function and $\cos$ denotes the cosine similarity.

Finally, we conduct contrastive learning on both phrase and aspect-opinion pair representations:
\begin{equation}
\small
    \begin{split}
        \mathcal{L}_{contra} &= \mathcal{L}_{contra}(\text{SP}^S, \text{SP}^T, t)\\
        &+\mathcal{L}_{contra}(\text{PAIR}^S, \text{PAIR}^T, t).
    \end{split}
    \label{eq:contar-all}
\end{equation}
$\text{SP}^S$ and $\text{SP}^T$ are the sets of phrase representations from the source and target domains. $\text{PAIR}^S$ and $\text{PAIR}^T$ are the sets of pair representations from the source and target domains.

\paragraph{Final Objective.}
We merge the ASTE loss of the source domain and contrastive loss in a joint way. The final training objective is:
\begin{equation}
    \mathcal{L} = \mathcal{L}_{aste} + \lambda \mathcal{L}_{contra},
    \label{eq:final-loss}
\end{equation}
where $\lambda$ is the hyper-parameter denoting the weight of contrastive loss.

\section{Experiments}
\begin{table*}[]
    \centering
    \small
    \begin{tabular}{cccccccc}
    \toprule
        Domain Pair & 14R$\to$14L & 15R$\to$14L & 16R$\to$14L & 14L$\to$14R & 14L$\to$15R & 14L$\to$16R & Average\\
    \midrule
        GTS             & 42.41$\pm$\scriptsize{0.75} & 33.02$\pm$\scriptsize{2.04} & 35.97$\pm$\scriptsize{1.15} & 49.64$\pm$\scriptsize{2.83} & 42.79$\pm$\scriptsize{3.10} & 47.21$\pm$\scriptsize{2.28} & 41.84\\
        BMRC            & 41.05$\pm$\scriptsize{2.07} & 32.21$\pm$\scriptsize{0.64} & 32.56$\pm$\scriptsize{2.30} & 54.41$\pm$\scriptsize{1.98} & 43.17$\pm$\scriptsize{1.74} & 52.46$\pm$\scriptsize{0.87} & 42.64\\
        BART-ABSA       & 45.38$\pm$\scriptsize{1.88} & 35.70$\pm$\scriptsize{3.21} & 37.87$\pm$\scriptsize{1.79} & 54.77$\pm$\scriptsize{1.29} & 47.82$\pm$\scriptsize{2.35} & 56.52$\pm$\scriptsize{2.07} & 46.34\\
        Span-ASTE & 44.79$\pm$\scriptsize{0.57} & 42.06$\pm$\scriptsize{0.72} & 41.43$\pm$\scriptsize{0.91} & 54.66$\pm$\scriptsize{1.69} & 46.86$\pm$\scriptsize{1.30} & 52.73$\pm$\scriptsize{1.39} & 47.09\\
    \midrule
        BMRC + AT       & 39.86$\pm$\scriptsize{2.14} & 31.89$\pm$\scriptsize{0.71} & 33.91$\pm$\scriptsize{2.10} & 55.06$\pm$\scriptsize{0.83} & 44.60$\pm$\scriptsize{1.61} & 50.78$\pm$\scriptsize{1.85} & 42.68\\
        Span-ASTE + AT  & 42.24$\pm$\scriptsize{1.02} & 41.10$\pm$\scriptsize{0.88} & 41.16$\pm$\scriptsize{1.03} & 51.73$\pm$\scriptsize{1.60} & 47.15$\pm$\scriptsize{2.12} & 48.95$\pm$\scriptsize{2.64} & 45.39\\
        \method         & \textbf{46.62}$\pm$\scriptsize{1.90} & \textbf{43.05}$\pm$\scriptsize{1.69} & \textbf{42.23}$\pm$\scriptsize{1.79} & \textbf{59.04}$\pm$\scriptsize{0.92} & \textbf{52.19}$\pm$\scriptsize{0.43} & \textbf{57.57}$\pm$\scriptsize{2.25} & \textbf{50.12}\\
    \bottomrule
    \end{tabular}
    \caption{F1 scores of different methods on six transfer pairs. R and L are the abbreviations for restaurant and laptop. We highlight the best results in bold.}
    \label{tab:main-result}
\end{table*}
\begin{table}[]
    \centering
    \small
    \begin{tabular}{lcccc}
        \toprule
        Method & \method & w/o phrase & w/o pair & w/o thres. \\
        \midrule
        F1 & 50.12 & 49.53 & 49.49 & 49.63\\
        \bottomrule
    \end{tabular}
    \caption{Ablation study results, where w/o phrase and w/o pair denote \method without phrase-level and pair-level contrastive learning and w/o thres. denotes \method without threshold in positive pair construction. All values are average F1 scores on  six transfer pairs.}
    \label{tab:ablation}
\end{table}

\subsection{Experimental Setup}
\paragraph{Datasets.} We evaluate \method on the ASTE dataset from \citet{xu-etal-2020-position}. It contains data from two domains, i.e., restaurant and laptop. We construct six transfer pairs based on the dataset and conduct cross-domain experiments on them. We train the model with labeled data from the source domain and unlabeled data from the target domain, then test the performance on the target domain.

\paragraph{Baselines.} We compare \method with several highly competitive ASTE methods: (1) BMRC \citep{chen2021bidirectional}: a machine reading comprehension (MRC) method with bidirectional queries. (2) BART-ABSA \citep{yan-etal-2021-unified}: a generative-based method with pointer network. (3) GTS \citep{wu-etal-2020-grid}: a grid tagging method that decode the triplet information from the a 2-dimension grid. (4) Span-ASTE \citep{xu-etal-2021-learning}: our backbone model that first extracts the aspect and opinion terms and then identifies the sentiment in an end-to-end way. Moreover, we incorporate some of these methods with domain adversarial training \citep[AT;][]{ganin2016domain}, a dominant solution in domain adaptation scenarios: (1) BMRC + AT: BMRC with adversarial training at the token level. (2) Span-ASTE + AT: Span-ASTE with adversarial training at the token level.

Due to space constraints, we present the implementation details in \textbf{Appendix} \ref{sec:app-exp-detail}.

\subsection{Main Results}
 We report the F1 scores of various methods in Table \ref{tab:main-result}. From the table, we make the following observations: 
(1) Span-ASTE achieves the best performance among baseline methods.
This motivates us to use Span-ASTE as the backbone model.
(2) Compared with Span-ASTE, \method achieves consistent gains over all transfer pairs, showing an improvement of 6\% on average.
These findings indicate that \method is a generally effective approach for cross-domain ASTE. 
(3) Adversarial training achieves minor or negative gains compared with baselines, which performance falls far behind \methodwb.
Overall, results demonstrate the effectiveness of \method for cross-domain ASTE.

\begin{table}[]
    \centering
    \small
    \begin{tabular}{llccccc}
    \toprule
        \multicolumn{2}{c}{Discrepancy}      & Span-ASTE   & \method & $\Delta$\\
    \midrule
        \multirow{2}{*}{Domain}      & phrase & 1.46 & 1.07 & -26.7\% \\
        ~                            & pair   & 2.00 & 1.75 & -12.5\% \\
    \midrule
        \multirow{2}{*}{Intra-class} & phrase & 1.92 & 1.60 & -16.7\% \\
        ~                            & pair   & 3.38 & 3.05 & -9.8\% \\
    \midrule
        \multirow{2}{*}{Inter-class} & phrase & 10.38 & 12.24 & +17.9\% \\
        ~                            & pair   & 6.03  & 7.30 & +21.1\% \\

    \bottomrule
    \end{tabular}
    \caption{Quantitative analysis for \methodwb. We demonstrate the domain discrepancy, intra-class and inter-class discrepancy for both phrases and aspect-opinion pairs.}
    \label{tab:quantitative}
\end{table}
\subsection{Ablation Study} 
To further evaluate the effectiveness of \methodwb, we conduct an ablation study on the following modules: (1) w/o phrase-level: \method with only aspect-opinion pair level contrastive loss, (2): w/o pair-level: \method with only phrase-level contrastive loss, (3) w/o threshold: \method without the sharpening threshold and features can be considered as positive pairs only if their highest predictions are of the same category. 
Results in Table \ref{tab:ablation} show that each component contributes to the performance of \methodwb. 

\subsection{Quantitative Analysis}
To evaluate the effectiveness of \method in improving transferability and preserving discriminability, we perform quantitative analysis on the Span-ASTE and \methodwb. Specifically, we employ trained models to obtain phrase and pair representations as shown in Equation \ref{eq:aste-rep}. Then we calculate the domain discrepancy, intra-class and inter-class discrepancy using Maximum Mean Discrepancy (MMD) as described in \citet{gretton2012kernel}, and present the results in Table \ref{tab:ablation}.
Results indicate that \method can reduce domain and intra-class discrepancy by 26.7\% and 16.7\%, respectively, while increasing inter-class discrepancy by 17.9\% for phrase representations. Similar results were obtained for pair representations. These findings suggest that \method improves transferability across domains while preserving discriminability in cross-domain ASTE task.


Due to space constraints, we provide hyper-parameter analysis and detailed experiment results for  adversarial training in \textbf{Appendix} \ref{sec:app-hyper} and \ref{sec:app-at}.
\section{Conclusion}
We propose \methodwb, a novel method for cross-domain ASTE. The method focuses on improving the transferability across domains and preserving the discriminability of different categories. Empirical experiments show that \method outperforms the baseline model by 6\% in the F1 score. 
\section*{Limitations}
The limitation of this study is the limited evaluation scenarios. In the experiments, we evaluate the performance when transferring from the restaurant domain to the laptop domain and from the laptop domain to the restaurant domain since restaurant and laptop are the only two domains available in existing datasets. Future works can validate the model performances on more diversified transfer pairs. 

\bibliography{acl2023}
\bibliographystyle{acl_natbib}

\appendix
\clearpage
\section{Related Work}
\label{sec:related}
Due to the scarcity of prior studies on cross-domain ASTE, this research presents a comprehensive review of previous works in the related areas of cross-domain sentiment analysis and aspect-based sentiment analysis. Furthermore, a discussion on contrastive learning is also included to further enhance the understanding of the field.

\paragraph{Cross-domain SA and Cross-domain ABSA.}
Most of the previous works on cross-domain sentiment analysis (SA) and aspect-based sentiment analysis (ABSA) can be separated into two groups: feature-based and data-based methods. The \textbf{feature-based} methods focus on learning domain-invariant features by leveraging auxiliary tasks \citep{yu-jiang-2017-leveraging, yang2019neural, zhang2021eatn} like sentiment detection  and using pivot feature to bridge the source and target domains  \citep{chernyshevich-2014-ihs, ziser-reichart-2018-pivot, wang-pan-2018-recursive, wang-pan-2019-syntactically}. The \textbf{data-based} methods aim to re-weighting the training data, that is, assigning higher weights to the reviews or words similar to the target domain and lower weights to those different from the target domain. \citet{li-etal-2012-cross} construct pseudo-labeled data in the target domain, and re-weight the source data based on the pseudo-labeled data. \citet{gong-etal-2020-unified} propose a unified framework and combine the feature and data-based methods. 

However, these researches all focus on sentence or aspect-level classification problems, which can not be directly adapted to the ASTE task. ASTE focuses on more fine-grained sentimental information and the sentiment triplets between the source and target domains are of huge differences. Therefore, we need a fine-grained domain adaptation strategy for cross-domain ASTE.

\paragraph{Contrastive Learning.} Contrastive learning has recently become a dominant solution  in self-supervised representation learning \citep{jaiswal2020survey}. It first constructs semantically similar positive pairs by data augmentation \citep{chen2020simple, misra2020self} and regards other instances in the dataset as negative examples. Then by pulling the positive pairs together and pushing the negative pairs away, it can learn semantics for the embedding. Motivated by studies in representation learning, we propose to apply contrastive learning to domain adaptation problems. By constructing positive and negative pairs from both the source and target domains, we can reduce the domain discrepancy from different categories, thereby improving the transferability and discriminability.

\section{Experiments}
\subsection{Implementation Details}
\label{sec:app-exp-detail}

\begin{table*}[]
    \small
    \centering
    \begin{tabular}{lcccc|cccc|cccc|cccc}
    \toprule
         \multirow{2}{*}{Domain} & \multicolumn{4}{c}{14Res} & \multicolumn{4}{c}{14Lap} & \multicolumn{4}{c}{15Res} & \multicolumn{4}{c}{16Res}   \\
         ~& \#S & \#+ & \#O &\#-& \#S & \#+ & \#O &\#-& \#S & \#+ & \#O &\#-& \#S & \#+ & \#O &\#-\\
    \midrule
         Train & 1266 & 1692 & 166 & 480 & 906 & 817 & 126 & 517 & 605 & 783 & 25 & 205 & 857 & 1015 & 50 & 329\\
         Dev   &310 & 404 & 54 & 119 & 219 & 169 & 36 & 141 & 148 & 185 & 11 & 53 & 210 & 252 & 11 & 76\\
         Test & 492 & 773 & 66 & 155 & 328 & 364 & 63 & 116 & 322 & 317 & 25 & 143 & 326 & 407 & 29 & 78\\
    \bottomrule
    \end{tabular}
    \caption{Statistics of \citet{xu-etal-2020-position}. \#S denotes the number of sentences. \#+, \#O, \#- denote the numbers of positive, neutral, and negative triplets, respectively.}
    \label{tab:statistics}
\end{table*}
\paragraph{Training Details.} We train the model with labeled data from the source domain and unlabeled data from the target domain in \citet{xu-etal-2020-position} (Table \ref{tab:statistics}). We run all the experiments five times with random seeds from 0 to 4 on NVIDIA A100 GPU with PyTorch. The pre-trained model is obtained from \href{https://huggingface.co/}{HuaggingFace}. We use AdamW optimizer \citep{loshchilov2017decoupled} for optimization. The learning rates for BERT and classifier are set to $5\cdot 10^{-5}$ and $1\cdot 10^{-3}$, respectively. We perform grid research for the hyper-parameters. All these hyper-parameters are tuned on the validation set. 

\paragraph{Evaluation Metric.} Following \citet{xu-etal-2021-learning}, we employ the F1 score to measure the performance of different approaches, where only extract matches can be considered correct.

\begin{figure*}
     \centering
     \begin{subfigure}[b]{0.32\textwidth}
         \centering
         \includegraphics[width=\textwidth]{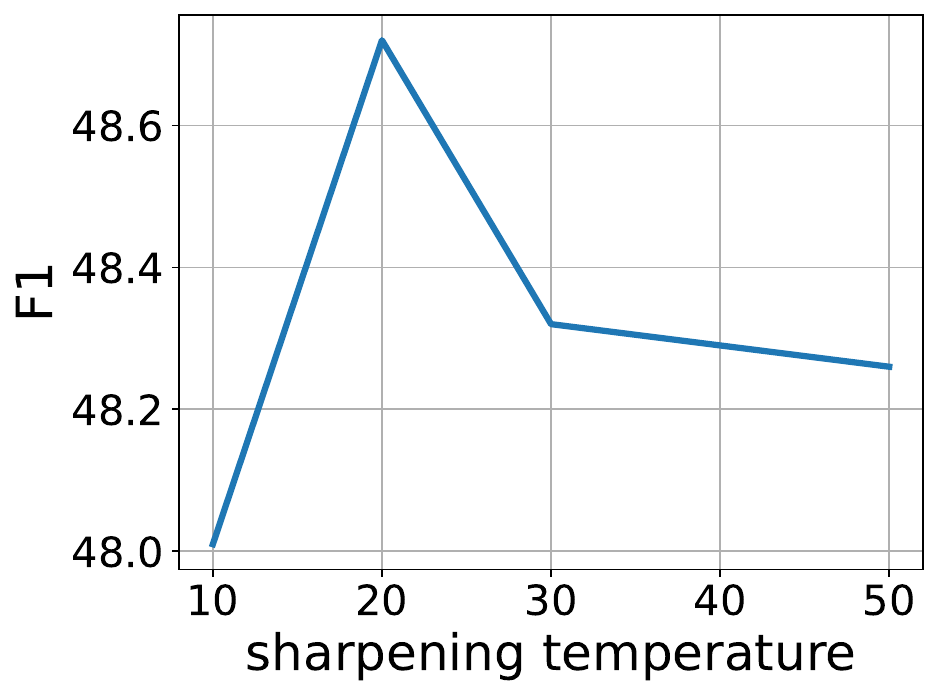}
     \end{subfigure}
        \hfill
     \begin{subfigure}[b]{0.32\textwidth}
         \centering
         \includegraphics[width=\textwidth]{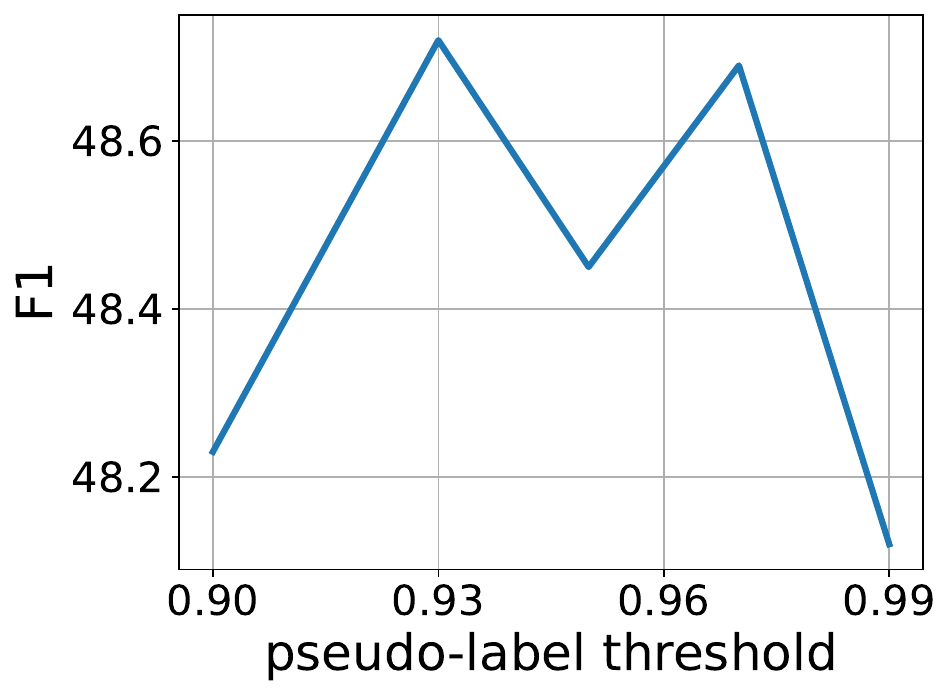}
     \end{subfigure}
     \hfill
     \begin{subfigure}[b]{0.32\textwidth}
         \centering
         \includegraphics[width=\textwidth]{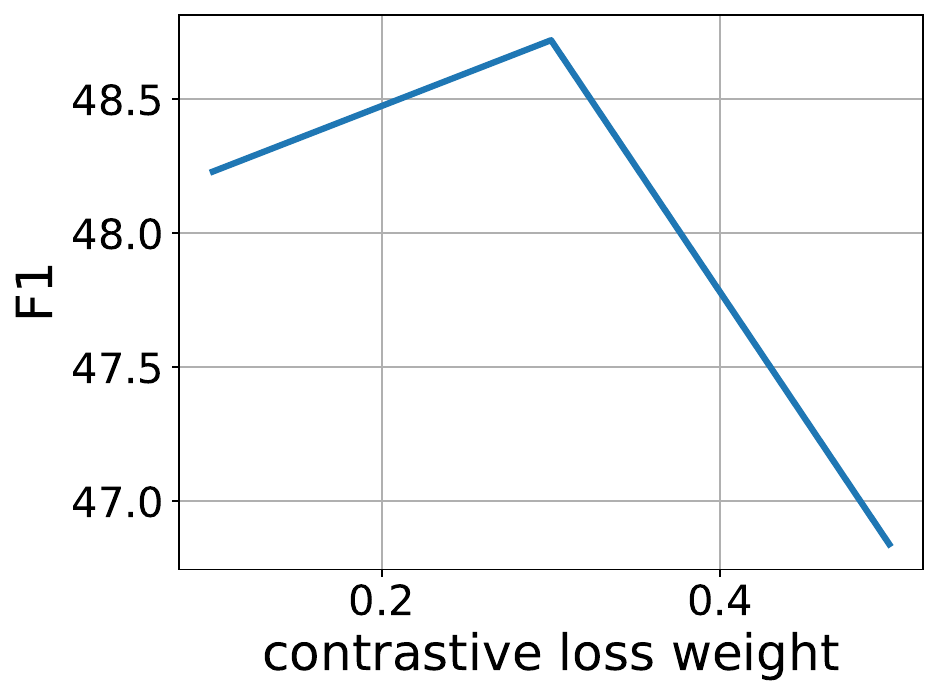}
     \end{subfigure}
        \caption{F1 scores on different sharpening temperature, pseudo-label threshold and contrastive loss weight.}
        \label{fig:hyper}
\end{figure*}

\subsection{Hyper-parameter Analyses}
\label{sec:app-hyper}
There are three hyper-parameters in \method: 
the sharpening temperature $\tau$, the pseudo-label threshold $t$, and the contrastive loss weight parameter $\lambda$. We tune all the hyper-parameters based on the model performance on the evaluation set. Results are shown in Figure \ref{fig:hyper}. We finally set $\tau=20, t=0.93, \lambda=0.3$ for all transfer pairs.

\subsection{Experiments of Adversarial Training}
\label{sec:app-at}
For adversarial training, we follow the implementation in \citet{ganin2016domain}. There is only one hyper-parameter for this method, $\alpha$, the ratio of training the generator to the discriminator. We search $\alpha$ in $\{1, 3, 5, 7, 10, 30, 50\}$ for Span-ASTE + AT and $\{1, 10, 50, 100, 500, 700, 1000, 1500\}$ for BMRC + AT. Then we select $\alpha$ based on the F1 score on the validation set. Finally, We set $\alpha=5$  for Span-ASTE + AT and $\alpha=1000$ for BMRC + AT. The parameter search costs about 1200 GPU hours.
Detailed experimental results are shown in Table \ref{tab:span-at} and Table \ref{tab:bmrc-at}. We can observe that adversarial training is unstable and parameter-sensitive. 
\begin{table*}[]
    \centering
    \small
    \begin{tabular}{cccccccc}
    \toprule
         $\alpha$ & 14res$\to$14lap & 15res$\to$14lap & 16res$\to$14lap & 14lap$\to$14res & 14lap$\to$15res & 14lap$\to$16res & Average\\
    \midrule
        1   & 32.33\scriptsize{$\pm$4.70} & 29.30\scriptsize{$\pm$5.62} & 32.95\scriptsize{$\pm$4.87} & 38.38\scriptsize{$\pm$11.70} & 42.73\scriptsize{$\pm$4.88} & 31.53\scriptsize{$\pm$13.39} & 34.54\\
        3   & 40.83\scriptsize{$\pm$1.35} & 35.44\scriptsize{$\pm$3.98} & 35.48\scriptsize{$\pm$7.79} & 45.61\scriptsize{$\pm$4.10} & 44.90\scriptsize{$\pm$2.93} & 50.47\scriptsize{$\pm$1.44} & 42.12\\
        5   & 39.40\scriptsize{$\pm$1.20} & 39.41\scriptsize{$\pm$1.14} & 38.74\scriptsize{$\pm$1.78} & 48.09\scriptsize{$\pm$2.27} & 52.14\scriptsize{$\pm$1.21} & 48.14\scriptsize{$\pm$3.43} & \textbf{44.32}\\
        7   & 38.65\scriptsize{$\pm$3.96} & 38.14\scriptsize{$\pm$1.49} & 37.34\scriptsize{$\pm$1.32} & 44.93\scriptsize{$\pm$1.77} & 47.92\scriptsize{$\pm$4.08} & 49.42\scriptsize{$\pm$1.43} & 42.73\\
        10  & 39.72\scriptsize{$\pm$1.85} & 38.95\scriptsize{$\pm$1.74} & 38.47\scriptsize{$\pm$1.21} & 46.81\scriptsize{$\pm$1.35} & 40.62\scriptsize{$\pm$20.58} & 47.19\scriptsize{$\pm$4.42} & 41.96\\
        30  & 40.87\scriptsize{$\pm$2.87} & 37.46\scriptsize{$\pm$1.48} & 38.00\scriptsize{$\pm$0.69} & 46.46\scriptsize{$\pm$5.00} & 49.10\scriptsize{$\pm$3.17} & 45.88\scriptsize{$\pm$3.88} & 42.96\\
        50  & 40.93\scriptsize{$\pm$1.80} & 36.97\scriptsize{$\pm$1.22} & 36.59\scriptsize{$\pm$1.34} & 47.79\scriptsize{$\pm$1.41} & 36.50\scriptsize{$\pm$18.32} & 47.32\scriptsize{$\pm$3.31} & 41.02\\
    \bottomrule
    \end{tabular}
    \caption{F1 scores of Span-ASTE + AT. All the results are reported on the evaluation set. We highlight the best average results in bold.}
    \label{tab:span-at}
\end{table*}
\begin{table*}[]
    \centering
    \small
    \begin{tabular}{cccccccc}
    \toprule
         $\alpha$ & 14res$\to$14lap & 15res$\to$14lap & 16res$\to$14lap & 14lap$\to$14res & 14lap$\to$15res & 14lap$\to$16res & Average\\
    \midrule
         1      & 33.43\scriptsize{$\pm$2.92} & 31.16\scriptsize{$\pm$3.02} & 30.15\scriptsize{$\pm$2.28} & 40.00\scriptsize{$\pm$2.71} & 38.17\scriptsize{$\pm$3.50} & 43.18\scriptsize{$\pm$2.99} & 36.02\\
         10     & 29.70\scriptsize{$\pm$4.49} & 31.36\scriptsize{$\pm$1.85} & 29.75\scriptsize{$\pm$4.53} & 34.14\scriptsize{$\pm$2.62} & 34.57\scriptsize{$\pm$4.02} & 40.26\scriptsize{$\pm$3.82} & 33.30\\
         50     & 37.47\scriptsize{$\pm$0.70} & 32.93\scriptsize{$\pm$1.22} & 32.26\scriptsize{$\pm$2.26} & 44.60\scriptsize{$\pm$1.72} & 49.66\scriptsize{$\pm$2.48} & 46.18\scriptsize{$\pm$4.40} & 40.52\\
         100    & 36.93\scriptsize{$\pm$2.13} & 34.96\scriptsize{$\pm$0.98} & 33.88\scriptsize{$\pm$2.12} & 45.02\scriptsize{$\pm$3.05} & 50.13\scriptsize{$\pm$2.86} & 49.31\scriptsize{$\pm$1.10} & 41.71\\
         500    & 38.27\scriptsize{$\pm$1.03} & 33.24\scriptsize{$\pm$0.85} & 33.06\scriptsize{$\pm$2.70} & 48.18\scriptsize{$\pm$1.17} & 50.48\scriptsize{$\pm$2.18} & 52.49\scriptsize{$\pm$1.48} & 42.62\\
         700    & 37.11\scriptsize{$\pm$1.97} & 34.34\scriptsize{$\pm$2.09} & 32.69\scriptsize{$\pm$2.32} & 47.15\scriptsize{$\pm$1.62} & 51.18\scriptsize{$\pm$1.36} & 52.02\scriptsize{$\pm$2.02} & 42.42\\
         1000   & 38.22\scriptsize{$\pm$1.81} & 33.09\scriptsize{$\pm$1.88} & 33.77\scriptsize{$\pm$2.07} & 48.29\scriptsize{$\pm$0.73} & 53.28\scriptsize{$\pm$1.14} & 52.12\scriptsize{$\pm$1.71} & \textbf{43.13}\\
         1500   & 37.49\scriptsize{$\pm$1.53} & 34.64\scriptsize{$\pm$0.37} & 33.11\scriptsize{$\pm$1.67} & 47.13\scriptsize{$\pm$3.21} & 51.07\scriptsize{$\pm$3.46} & 51.14\scriptsize{$\pm$2.26} & 42.43\\
    \bottomrule
    \end{tabular}
    \caption{F1 scores of BMRC + AT. All the results are reported on the evaluation set. We highlight the best average results in bold.}
    \label{tab:bmrc-at}
\end{table*}

\end{document}